%% file: main.tex
\documentclass[10pt,twocolumn,letterpaper]{article}
\usepackage[accsupp]{axessibility} 
\usepackage[pagenumbers]{cvpr} 

\input{preamble}

\definecolor{cvprblue}{rgb}{0.21,0.49,0.74}
\usepackage[pagebackref,breaklinks,colorlinks,citecolor=cvprblue]{hyperref}

\def\paperID{14461} 
\def\confName{CVPR}
\def\confYear{2024}

\title{Fooling Polarization-based Vision using Locally Controllable Polarizing Projection}

\author{Zhuoxiao Li$^1$ \quad Zhihang Zhong$^2$ \quad Shohei Nobuhara$^3$ \quad Ko Nishino$^3$ \quad Yinqiang Zheng$^1$\thanks{Corresponding author}\\
$^1$The University of Tokyo \quad $^2$Shanghai Artificial Intelligence Laboratory \quad $^3$Kyoto University
}

\begin{document}
\maketitle
\input{sec/0_abstract}
\input{sec/1_intro}
\input{sec/2_related}
\input{sec/3_preliminary}
\input{sec/4_glass}
\input{sec/5_shape}

\input{sec/6_conclusion}
{
    \small
    \bibliographystyle{ieeenat_fullname}
    \bibliography{main}
}


\end{document}

%% file: preamble.tex
\usepackage[dvipsnames]{xcolor}


%% file: sec/0_abstract.tex
\begin{abstract}
Polarization is a fundamental property of light that encodes abundant information regarding surface shape, material, illumination and viewing geometry. The computer vision community has witnessed a blossom of polarization-based vision applications, such as reflection removal, shape-from-polarization (SfP), transparent object segmentation and color constancy, partially due to the emergence of single-chip mono/color polarization sensors that make polarization data acquisition easier than ever. However, is polarization-based vision vulnerable to adversarial attacks? If so, is that possible to realize these adversarial attacks in the physical world, without being perceived by human eyes?  In this paper, we warn the community of the vulnerability of polarization-based vision, which can be more serious than RGB-based vision. By adapting a commercial LCD projector, we achieve locally controllable polarizing projection, which is successfully utilized to fool state-of-the-art polarization-based vision algorithms for glass segmentation and SfP. Compared with existing physical attacks on RGB-based vision, which always suffer from the trade-off between attack efficacy and eye conceivability, the adversarial attackers based on polarizing projection are contact-free and visually imperceptible, since naked human eyes can rarely perceive the difference of viciously manipulated polarizing light and ordinary illumination. This poses unprecedented risks on polarization-based vision, for which due attentions should be paid and counter measures be considered. 
\end{abstract}

%% file: sec/1_intro.tex
\section{Introduction}
\label{sec:intro}

\begin{figure}[t]
	\centering
	\includegraphics[width=0.98\linewidth]{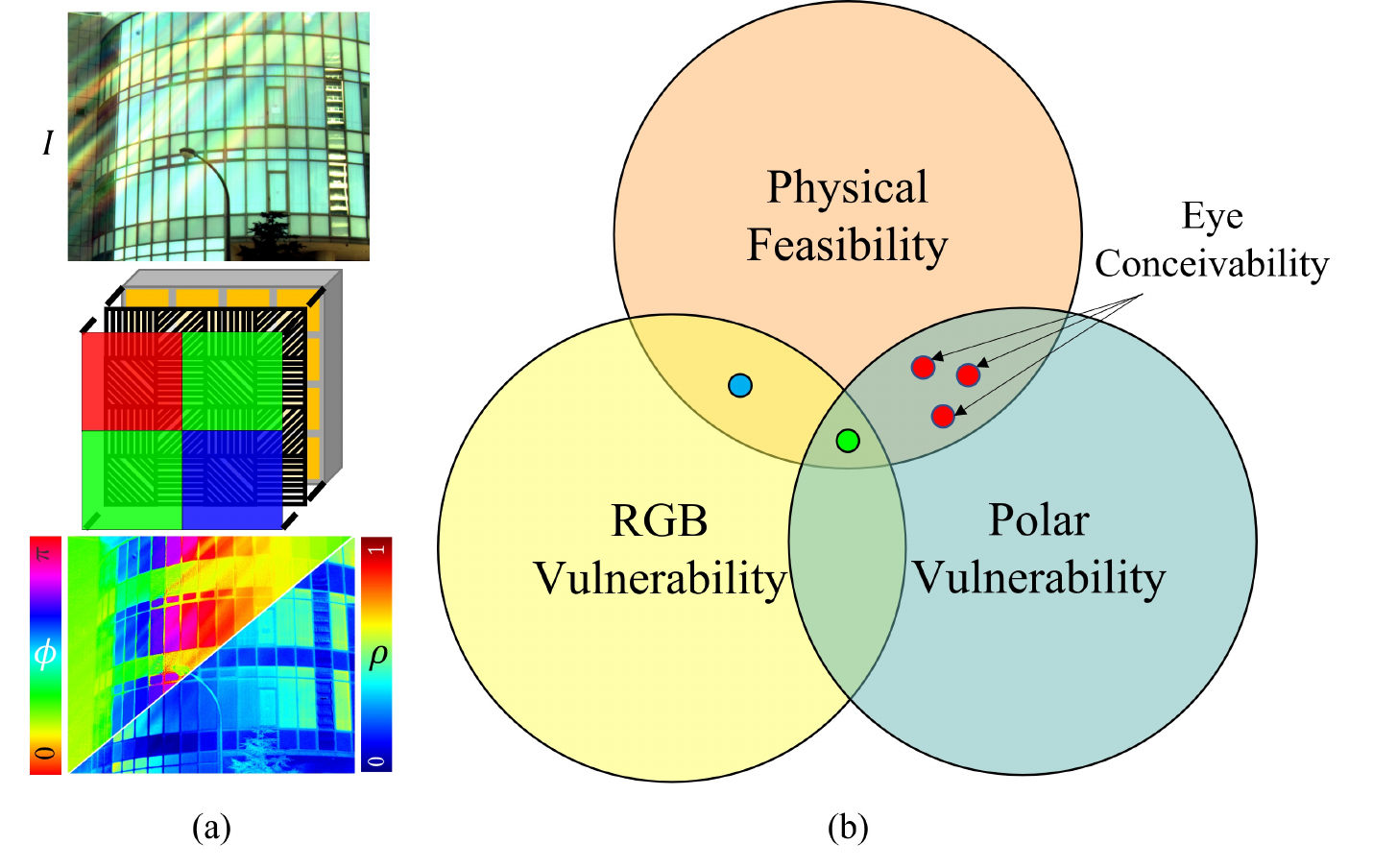}
	\caption{(a) Single-chip color polarization sensor can capture trichromatic image, angle of linear polarization (AoLP), and degree of linear polarization (DoLP) within one shot. (b) Our proposed physical attackers are based on polarizing projection, which is naturally conceivable to human eyes, thus can bypass the trade-off between attack efficacy and eye conceivability in fooling RGB-based vision.}
	\label{fig:first}
\end{figure}

Even if the frequency of light lies in the visible range, its polarization status can hardly be perceived by human eyes. Fortunately, a variety of imaging devices have been developed, which allow to utilize rich scene information encoded in polarization, regarding geometry, material, illumination and light transportation. The emergence of single-chip mono/color polarization sensors has made polarization data acquisition easier, leading to a blossom of polarization-based vision applications, such as reflection removal~\cite{lei2020polarized}, shape-from-polarization (SfP)~\cite{deschaintre2021deep, fukao2021polarimetric, lei2022shape}, surface defects detection~\cite{li2021multisensor}, color constancy~\cite{ono2022}, transparent object detection and segmentation~\cite{mei2022glass}. Figure~\ref{fig:first} (a) shows the capabilities of a single-chip color polarization sensor in capturing trichromatic image $I$, angle of linear polarization (AoLP) $\phi$, and degree of linear polarization (DoLP) $\rho$, with one shot. Given the prevalence of polarization-based vision, it is astonishing that its vulnerability has never been formally explored in the CV and AI communities. 

The vulnerability of RGB-based deep vision models is firstly reported in~\cite{szegedy2013intriguing}, with various extensions in minimizing perturbation magnitude, maximizing success rate of attack, retrieving universal adversarial attackers, and so on~\cite{Akhtar2018survery,Akhtar2021survery}. Stepping beyond the digital space, more recent researches focus on studying the vulnerability of RGB-based vision models in the physically feasible space, while minimizing the level of offensiveness to human eyes (blue point in Figure~\ref{fig:first}(b)), by using printed attack patterns on papers~\cite{kurakin2018adversarial, athalye2018synthesizing, jan2019connecting}, clothes~\cite{wu2020making}, or perturbations projected by projectors~\cite{gnanasambandam2021optical}, and attacks created by laser beams~\cite{duan2021adversarial} and shadows~\cite{zhong2022shadows}. In principle, physical adversarial attack can be more catastrophic, since there is no need to hack the input image or the deployed model as digital attack requires. However, it is extremely hard to find a physical attacker that is both effective and imperceptible, since RGB cameras by design are mimicking human eyes, and a physically feasible attack that is invisible to eyes will not be captured by the camera as well. 

Given that polarization and color represent two distinct dimensions of light, and polarization is usually introduced as a complementary modality to assist RGB-based vision, one might believe that, polarization-based vision, especially when coupled with the RGB modality, should be safer and harder to be attacked in the physical world, as the green dot in Figure~\ref{fig:first}(b) illustrates. In this paper, we show that this speculation is ungrounded by proposing a novel yet simple implementation of locally controllable polarizing projection. Since human eyes have no sensitivity to polarization, the most stringent restriction on eye conceivability in attacking RGB-based vision is naturally bypassed. This allows us to explore the vulnerability space of polarization-based vision more flexibly, within in the broad feasible space that the projector can realize (red dots in Figure~\ref{fig:first}(b)). 

Inspired by the operating principle of Liquid Crystal Display (LCD) panels in monitors and projectors, we have recognized that the polarization status of light emitted from each liquid crystal cell can be independently controlled, after removing the front polarization film attached onto the LCD panel. Since human eyes can not perceive polarization status, the projected light looks uniformly white, even if the projection pattern has colors and textures, and the polarization status of light has been adjusted accordingly by LCD. In contrast, polarization cameras can record the programmed polarizing projection, and the behaviors of vision algorithms based on such information might be manipulated. 

We have verified the feasibility of fooling polarization-based vision for two representative tasks  via whitebox attack, including (i) reducing the accuracy of RGB-polar-based glass segmentation~\cite{mei2022glass}; (ii) misleading the latest shape estimation model relying on polarization~\cite{lei2022shape}. We hope this study can arouse attention on the potential security risks of utilizing polarization and trigger further researches on the defense side.

%% file: sec/2_related.tex
\section{Related Work}
\label{sec:related}

\subsection{Adversarial Attacks on RGB-based Vision}
While the state-of-the-art deep neural networks are capable of achieving incredible performance in various scene understanding tasks, recent researches~\cite{kurakin2018adversarial, szegedy2013intriguing} revealed their striking vulnerability that very mild modifications to the input images can deceive advanced classifiers with high confidence.  
The adversarial examples are generated through optimization processes by maximizing the classification error of a targeted model. In digital world, on the premise of direct access to the targeted model, an adversarial example can be derived by one or multiple steps of perturbation following negative gradient directions, including classic Fast Gradient Sign Method (FGSM)~\cite{goodfellow2014explaining}, the Basic Iterative Method (BIM)~\cite{goodfellow2014explaining}, and the Projected Gradient Descent (PGD)~\cite{madry2017towards} for efficient and transferable adversarial attacks. 
Their perturbations are bounded with a small norm-ball $L_p<\epsilon$, normally $p=2$ or $\infty$, or minimized with a joint adversarial loss~\cite{carlini2017towards}, to craft a quasi-imperceptible example to human eyes. 

Digital attacks assume they can hijack the prediction system to directly feed adversarial examples into the targeted model. Considering that this requirement is usually impractical, other researches try to realize adversarial attacks by inserting perturbations into the physical world. \cite{kurakin2018adversarial} shows adversarial examples printed on papers are partly effective to fool DNN classifiers. However, because of the discrepancy between the designed attacker in the digital space and the physical attacker recorded by the camera, a key task is to retrieve robust adversarial examples that can be faithfully realized. \cite{jan2019connecting} approximates the full digital-to-physical transformation to search perturbations in a simulated world. To deal with the wide range of diversities in real world scenarios, e.g. view points, illuminations, and noises, \cite{athalye2018synthesizing} gets a distribution of transformations involved in the optimization procedure, including rescaling, rotation (in 2D or 3D), translation of image, and so on. 

However, former small perturbations are too subtle to be captured by cameras in the wild. Therefore, recent physical-world adversarial attack methods attempt to generate strong but stealthy perturbations in the real world. For example, stickers and graffiti-type perturbations are attached to targeted objects, e.g. a road sign, to achieve targeted misclassification from arbitrary viewpoints. Wearable attack perturbations like clothes~\cite{xu2020adversarial} and eye-glasses~\cite{sharif2016accessorize} are capable of fooling detection systems with improved stealthiness. Moreover, laser beams~\cite{duan2021adversarial}, shadows~\cite{zhong2022shadows}, and projection~\cite{gnanasambandam2021optical} are utilized to craft attack perturbations in the physical world without touching target objects. We refer readers to~\cite{wei2022survery} for thorough literature reviews on physical adversarial attacks. All these researches on physical adversarial attacks have to make a trade-off between attack efficacy and eye conceivability.

\subsection{Polarization-based Vision}
Polarization has been utilized in various vision tasks for many years, which is further boosted recently due to the emergence of a single-chip polarimetric imaging sensor that provides chromatic and polarimetric information in a single shot. Polarization cues are highly related to the illumination condition, object geometry and material property. Thus, it is introduced to assist conventional stereo vision approaches, i.e. multi-view stereo~\cite{cui2017polarimetric, yang2018polarimetric}, binocular stereo~\cite{fukao2021polarimetric}, and photometric stereo~\cite{tozza2021uncalibrated}. Besides, polarized specular and diffuse reflection have distinct properties, which make it also popular in inverse rendering by formulating the physics-based rendering equation with polarimetric Bidirectional Reflection Distribution Function (pBRDF) model~\cite{baek2018simultaneous, hwang2022sparse, zhao2022polarimetric}. With learned shading and polarization priors, DNNs can restore more detailed geometries~\cite{ba2020deep, lei2022shape} and SVBRDF~\cite{deschaintre2021deep} with a single shot. Moreover, polarimetric imaging is capable of capturing polarization cues of transparent objects, which explains its extraordinary superiority in dealing with transparent objects in glass segmentation~\cite{mei2022glass}, transparent object shape estimation~\cite{mingqi2022transparent}, and reflection removal~\cite{lei2020polarized}. Color constancy is challenging in the RGB domain, and it is shown that polarization can benefit color constancy, especially in poorly illuminated conditions~\cite{ono2022}. 

Wider applications of polarization in the near future can be expected, yet we would like to warn of the potential vulnerability of polarization-based vision, which might be more serious than that of RGB-based vision, since the adversarial attackers can be physically realized using a modified LCD projector and human eyes can not differentiate maliciously manipulated polarizing light from normal illumination. 

\subsection{Projectors and Their Applications}
Projectors are widespread display devices, whose modulation mechanism of light intensity is either based on digital micromirror device (DMD) widely used in digital light processing (DLP) projectors or liquid crystal polarization adopted by LCD projectors or LCoS projectors. As for the color-framing mechanism, one-chip DLP projectors use the rotating color wheel or blinking trichromatic LEDs, and one-chip LCD projectors use the micro color filter array, which is similar to the Bayer pattern in RGB cameras. By using the color-framing mechanism of a one-chip DLP projector, Ashdown et al.~\cite{ashdown2006robust} recovered high-resolution spectral reflectance. Further, to deal with unexpected irregularities when applying a digital projector in non-ideal situations, they proposed to generate a compensation image based on both the radiometric model of the system and the content of the image.
Tanaka et al.~\cite{tanaka2015recovering} utilized a projector coaxially placed with the camera to inject illuminations of multiple frequencies for obtaining the appearance of individual inner slices. Projectors have also been used for adversarial attacks in the physical world with projected perturbations~\cite{gnanasambandam2021optical} or constant colors~\cite{hu2022adversarial}. 

Existing polarization-based LCD/LCoS projectors do not offer pixel-wise manipulation of the polarization status of light projected on the screen. So, they can not be directly utilized to attack polarization-based vision algorithms. In the following, we will show that a simple adaptation of the one-chip LCD projector will allow locally controllable polarizing projection. 

%% file: sec/3_preliminary.tex
\section{Preliminary}

\begin{figure}[t]
	\centering
	\includegraphics[width=0.75\linewidth]{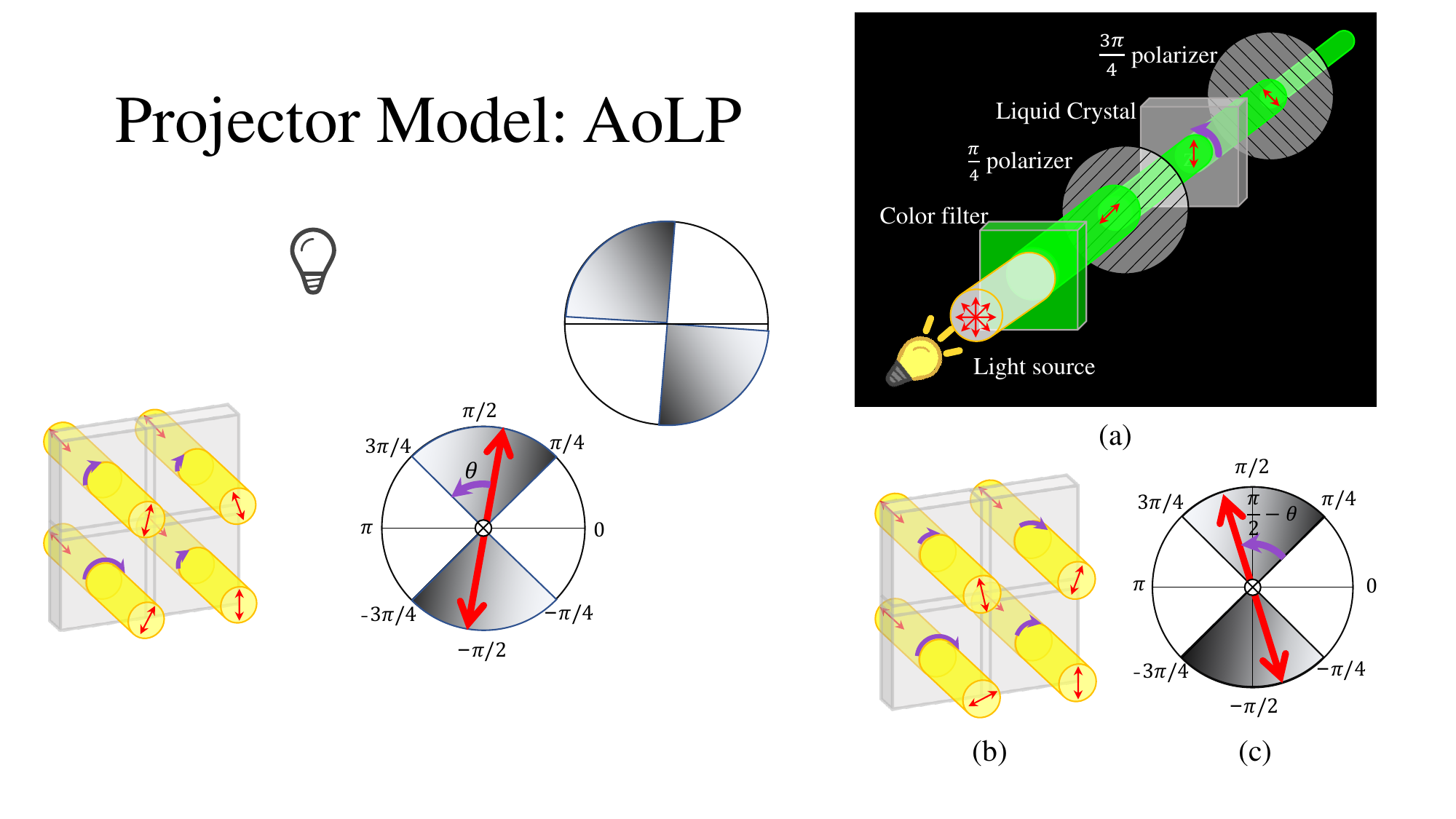}
	\caption{(a) The principle of intensity adjustment in one-chip LCD projector with a liquid crystal panel sandwiched by two perpendicular linear polarizers. (b) The polarization direction of the light beam in each liquid crystal cell can be individually controlled, without affecting its intensity. (c) The range of controllable polarization direction.}
	\label{fig:lcd}
	
\end{figure}

\begin{figure}[t]
	\centering
	\includegraphics[width=0.88\linewidth]{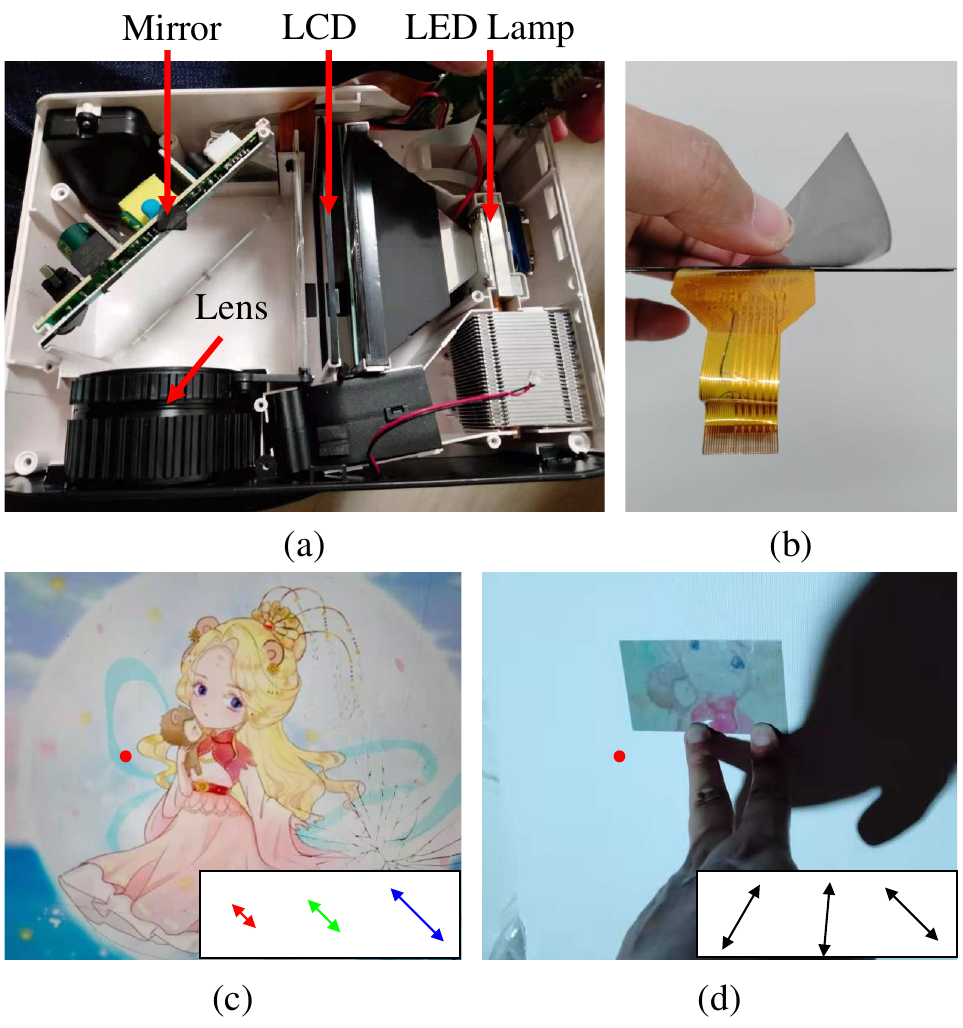}
	\caption{(a) The typical structure of a one-chip LCD projector, in which light emitted by LED lamp will go through a liquid crystal panel, mirror, and projection lens. (b) A linear polarizer is attached to the front side of the LCD panel. We tear it off to make our polarizing projection. (c)(d) The projection of a normal LCD projector and our adapted projector. The arrows' direction and length represent light polarizing direction and intensity, respectively. A normal projector emits out colorful light of constant polarization direction, while our adapted polarizing projector emits light with constant intensity but different polarizing angles. Note that, for naked eyes and ordinary RGB cameras, the projected light is completely uniform, even if their polarization directions are totally different. The projected image can be observed by eyes with the assistance of a linear polarizer on the screen.}
	\label{fig:reallcd}
\end{figure}

\subsection{Principle of One-chip LCD Projector}
One-chip LCD projector is the most widely used type of low-cost projector. The principle of an LCD projector controlling the irradiance is shown in Figure \ref{fig:lcd}. A beam of unpolarized light is emitted by a bulb. Two linear polarizers are placed in coaxial positions while their polarizing directions are perpendicular to each other ($\frac{\pi}{4}$ and $\frac{3\pi}{4}$ in our device), and a liquid crystal panel is inserted to the middle of them. The light is divided into red, green, and blue components by a color filter array before linearly polarized by the back $\frac{\pi}{4}$ polarizer. Then, by adding voltages to liquid crystal grids, the layer can manipulate polarizing direction of individual light beams, and a greater voltage leads to a bigger rotation up to $\frac{\pi}{2}$ from its initial direction. The light intensity passing through the front polarizer is decided by its polarizing direction, following the Malus's law:
\begin{equation}
	I=I_0\cos^2(\theta),
\end{equation}
where $\theta$ is the angle between the polarizing direction of light after being rotated by the liquid crystal and the direction of the front $\frac{3\pi}{4}$ polarizer. $I_0$ is a constant light intensity from the back polarizer. Since the intensity of RGB components is separately controlled, the color of merged light can be manipulated to match the projection pattern sufficiently. 

Note that, with the front polarizer equipped, output light beams are always linearly polarized in the $\frac{\pi}{4}$ axis related to the projector but have different intensities, as shown in Figure \ref{fig:reallcd} (c).  {\itshape The key idea of building a controllable polarization light projector is to remove the front polarizer}. As shown in Figure \ref{fig:reallcd} (a,b), we tear off the front polarization film of the projector and manage not to damage the liquid crystal panel. In this way, the output lights of the projector have constant intensity but different polarizing directions. The polarizing direction can be precisely controlled by manipulating the projection pattern. Furthermore, uniform white color and constant projection intensity contribute to high stealthiness as it will not introduce visible textures to human eyes, as can be seen in Figure \ref{fig:reallcd} (d).

\subsection{Preliminaries for Light Polarization}

Most polar-RGB based methods rely on both intensity and polarization cues, i.e., degree of linear polarization (DoLP, $\rho$, the proportion of linear polarized component in light) and angle of linear polarization (AoLP, $\phi$, the polarizing direction of polarized light). They can be calculated from a single shot with a Bayer-polarization sensor, \eg, IMX250MYR, which captures polarization components in four directions, termed as $I_{0}$, $I_{\frac{\pi}{4}}$, $I_{\frac{\pi}{2}}$, and $I_{\frac{3\pi}{4}}$. Stokes parameter, $\mathbf{s}=[s_0, s_1, s_2]^\top$ is used to describe the polarization state of light, where $s_0$ represents the total intensity of light, $s_1$ and $s_2$ describe the polarization states in horizontal and diagonal axes. $s_0$, $s_1$ and $s_2$ can be computed following:
\begin{equation}
	\begin{aligned}	
		& s_0 = (I_{0} + I_{\frac{\pi}{4}} + I_{\frac{\pi}{2}} + I_{\frac{3\pi}{4}})/2, \\
		& s_1 = I_{0} - I_{\frac{\pi}{2}}, \\
		& s_2 = I_{\frac{\pi}{4}} - I_{\frac{3\pi}{4}}.
	\end{aligned}
	\label{equ:i2s}
\end{equation}
Note that the integration of multiple light can be calculated as linear combination of their Stokes parameters. Then, $\rho$ and $\phi$ are generated by Stokes elements as:
\begin{equation}
	\begin{aligned}
		\rho = \frac{\sqrt{s_1^2 + s_2^2}}{s_0}, \phi = \frac{1}{2}\arctan\frac{s_2}{s_1}.
	\end{aligned}
	\label{equ:s2p}
\end{equation}
Also, $s_1$ and $s_2$ can be computed from $s_0$, $\rho$ and $\phi$ by:
\begin{equation}
	\begin{aligned}
		s_1=s_0\rho \cos(2\phi), s_2 = s_0\rho \sin(2\phi).
	\end{aligned}
	\label{equ:p2s}
\end{equation}

%% file: sec/4_glass.tex
\section{Whitebox Attack on Glass Segmentation}


Based on the novel locally controllable polarizing projection, we will show how to attack a polar-RGB based deep model, PGSNet for glass segmentation~\cite{mei2022glass}, in a whitebox manner. Assuming full access to the target model, our goal is to design an effective attack setting, find a stable perturbation, and project it onto the targeted scene in the physical world. The projected adversarial attacker will not be recognized by human eyes, since it appears to be uniformly white, yet can be captured by a polar-RGB camera.  Under such manipulated inputs, the PGSNet model is induced to generate incorrect predictions.

\subsection{Our Setting's Challenges}
Fine-scaled perturbations like pixel-wise noises are extremely subtle and easy to be destroyed in a complicated physical world. Therefore, previous works apply large perturbation patterns, like blocks~\cite{eykholt2018robust}, lines~\cite{duan2021adversarial}, triangles~\cite{zhong2022shadows}, for attacking image classification models. However, segmentation models predict pixel-wise classifications, which also apply advanced multiscale architectures, skip connections, and even self-attention modules, making them robust to adversarial attacks~\cite{arnab2018robustness,kamann2020benchmarking}. Our polarization projection involves complex physical and optical properties, which makes the creation of adversarial samples highly complex and technical. Thus, we develop a perturbation pattern in the form of grids, to realize a more robust physical attack.

Normally, whitebox adversarial attacks in the physical world need to simulate comprehensive effects in the real world, including camera response function, camera noises, light decay, quantization effects, and so on~\cite{gnanasambandam2021optical,jan2019connecting}. Therefore, in order to attack a polarization-based AI model, the most intuitive way of generating an effective perturbation in the whitebox manner is to simulate the complete transport of light. For us, polarization light travels from the polarizing projector to the object's surface, and finally to the polarization camera after being reflected by the object. However, it is not possible to acquire detailed scene geometry and accurate material parameters in the wild. So, we consider a simplified setup for our whitebox attacks. We can not only construct the most reliable simulation for polarization projection on glasses in the digital world but also generate adversarial examples of high effectiveness in the physical world.

\subsection{Digital World Attack}
Given a clean input $\mathbf{s}_{b}$ in the form of Stokes parameters and its binary label $y\in \mathbb{R}^{H\times W \times 1}$, where $(H, W)$ denotes spatial resolution. PGSNet $f(\cdot)$ is trained to maximize the pixel-wise binary prediction accuracy, where $1/0$ represents the region that is/is not glass. The goal of our adversarial attack is to maximize the segmentation error with a projected adversarial perturbation, denoted as $v$, which is captured as $\mathbf{s}=\mathbf{s}(v)$.
The problem is formulated as 
\begin{equation}
	\max_{v} \ \frac{1}{H W}\parallel y- f(g(\mathbf{s}_{b}+\mathbf{s}(v)))\parallel,
	\label{equ:p}
\end{equation}
where $g(\cdot)$ denotes mapping from Stokes vectors to polarization cues by equation \ref{equ:s2p}.

In general, adversarial attack algorithms generate adversarial examples by adding the gradient of error function w.r.t. $\mathbf{s}_b$, termed as $\nabla_{\mathbf{s}_b}\mathcal{L}$, and the perturbation is the division of clean image and its adversarial optimization result. However, the perturbation generated in this approach does not obey the physical property of our polarizing projection. For an optical adversarial system, we need to update perturbation directly following the gradient of the perturbation~\cite{gnanasambandam2021optical}, i.e., the projection pattern $v$ that will be fed into polarizing the projector, denoted as $\nabla_v\mathcal{L}$. 

However, as mentioned, it is impossible to accurately construct a differentiable computation of $\mathbf{s}(v)$, since the complicated reflection effects. Thus, we try to generate perturbation by directly optimizing $\mathbf{s}$. We generate our adversarial example from a collection of real-world polarization images, termed as $\mathbf{S}_p=\{\mathbf{s}_1, \mathbf{s}_2, ..., \mathbf{s}_K\}$. They are captured directly by the targeted camera in the scene covered by the corresponding uniform polarizing projections $\mathbf{v}_p=\{v_1, v_2, ..., v_K\}$. For robust attacks in the real world, our projected perturbation is a map of grids, the value of each grid is assigned with a selected value, \eg, $v_{i}$. With the relationship between the reflection $\mathbf{s}_i$ and the quantized projection $v_i$ known, the function $\mathbf{s}(v)$ is simply realized using a direct mapping. Then, we can  realize optimization using $\mathbf{S}_p$ rather than $\mathbf{v}_p$ to avoid complicated simulation of polarization reflection, indirect light effects, and the tone-mapping of projector. 
Then, we introduce a set of optimizable weights $\mathbf{\Omega}=\{\omega_1, \omega_2,...,\omega_K\}$ on candidate images and use $SoftMax$ function to generate relative coefficients of each $\mathbf{s}_i$. We compute an adversarial example as:
\begin{equation}
	\mathbf{s}_{ae} = \sum_i^K \frac{\exp({\omega_i/\tau})}{\sum_{j}^K \exp({\omega_j/\tau})} (\mathbf{s}_i - \mathbf{s}_b) + \mathbf{s}^*_b,
	\label{equ:sum}
\end{equation}
where $\tau$ is a temperature parameter to adjust the bias of relative weights. $\mathbf{s}^*_b$ denotes the augmented background image. Our optimization variables are the weights $\mathbf{\Omega}$. The problem in equation $\ref{equ:p}$ is then reformulated as:
\begin{equation}
	\max_\mathbf{\Omega}\frac{1}{HW} \parallel y- f(g(\mathbf{s}_{ae}))\parallel.
	\label{equ:aaa}
\end{equation}

To deal with the problem in equation \ref{equ:aaa}, we follow the negative gradient directions to update $\mathbf{\Omega}$ based on an iterative optimization approach:
\begin{equation}
	\mathbf{\Omega}^{t+1}  \leftarrow \mathbf{\Omega}^t+\alpha\nabla_{\mathbf{\Omega}^t}\mathcal{L}(y, f(g(\mathbf{s}_{ae}))),
	\label{equ:op}
\end{equation}
where $\alpha$ denotes the step size. After the optimization, we use $ArgMax$ to decide the final $\hat{\mathbf{\Omega}}$ and form an adversarial perturbation for a physical world attack. In practical terms, to strike a balance between efficiency and effectiveness, we assembled a set of 17 candidate images. These images have source projection values that are uniformly discrete, ranging from 0 to 255. Further implementation details can be found in the supplementary material.

\subsection{Adversarial Loss}
With our adversarial example, we aim to maximize the error between the predicted glass segmentation result $y_{ae}$ and label $y$, thus we first apply a Binary Cross Entropy loss. Moreover, we prefer to mislead the PGSNet to predict more non-glass pixels as positive, and vice versa, which can be seen as a targeted attack~\cite{gu2022segpgd}. With the prediction $y_{ae}$ of adversarial example, $\mathcal{L}_E$ is termed as:
\begin{equation}
	\begin{aligned}
		\mathcal{L}_E& = \frac{1}{HW}\sum_{j\in y^n} y_{ae}^j - \frac{1}{HW}\sum_{j\in y^p} y_{ae}^j,
		\label{equ:op}
	\end{aligned}
\end{equation}
where $y^{l\in{\{n, p\}}}$ denotes the set of pixels which are negative/positive (non-glass/glass) in the label $y$. $y^n$ and $y^p$ are disjoint, and the total number of pixels of $y^n \cup y^p$ is $HW$. 
Thus our final adversarial loss function is $\mathcal{L} = \mathcal{L}_{BCE} + \lambda\mathcal{L}_E$.
\subsection{Augmentation for Attack in Real World}
To generate more robust adversarial examples for the physical world attack, we follow the data augmentation strategy of EOT (Expectation Over Transformation)~\cite{athalye2018synthesizing}. EOT employs a distribution of real-world degradations and transformations, enabling the generation of adversarial examples that are better suited for the complexities of the physical world. Given our specific focus on a scenario with a fixed, known camera and projector setup, transformations such as rotation and translation are not applicable in our case. We introduce Gaussian noise and apply Gaussian blur to simulate real-world degradation. Additionally, we employ a randomly sampled scale ratio to adjust the intensity of the background image $\mathbf{s}_b$, accounting for minor variations in environmental lighting conditions.

\begin{figure}[t]
	\centering
	\includegraphics[width=1\linewidth]{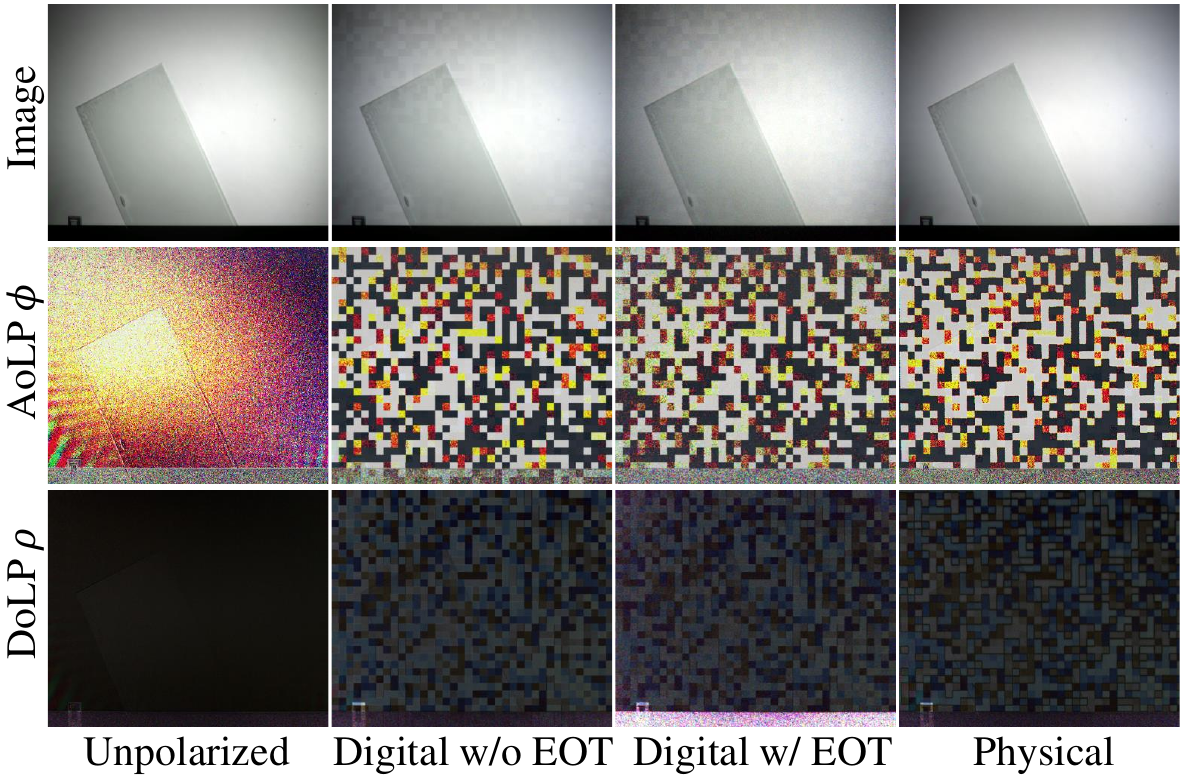}
	\caption{The Illustration of intensity, AoLP and DoLP images from our digital world simulation alongside their counterparts captured in the physical world. In the digital scenario with Expectation Over Transformation (EOT), the images are augmented through the addition of Gaussian noises, Gaussian blurring, and scaling of background intensity, all of which are randomly sampled to simulate real world degradations.}
	\label{fig:sim}
	
	\vspace{-.2cm}
\end{figure}

\begin{figure}[t]
	\centering
	\includegraphics[width=1\linewidth]{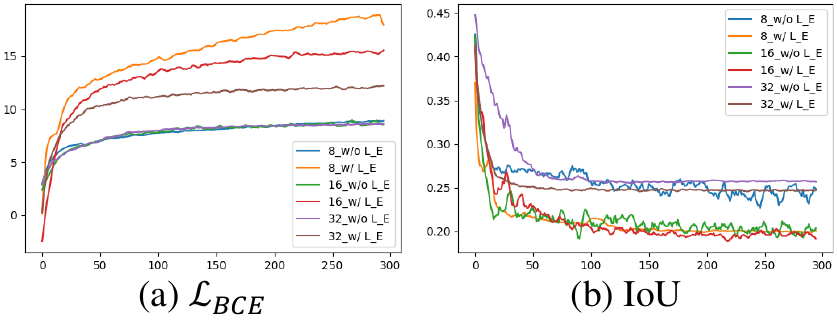}
	\caption{Illustration of $\mathcal{L}_{BCE}$, and IoU during the optimization process.}
	\label{fig:curve}
	\vspace{-.4cm}
\end{figure}

\begin{figure*}[t]
	\centering
	\includegraphics[width=1\linewidth]{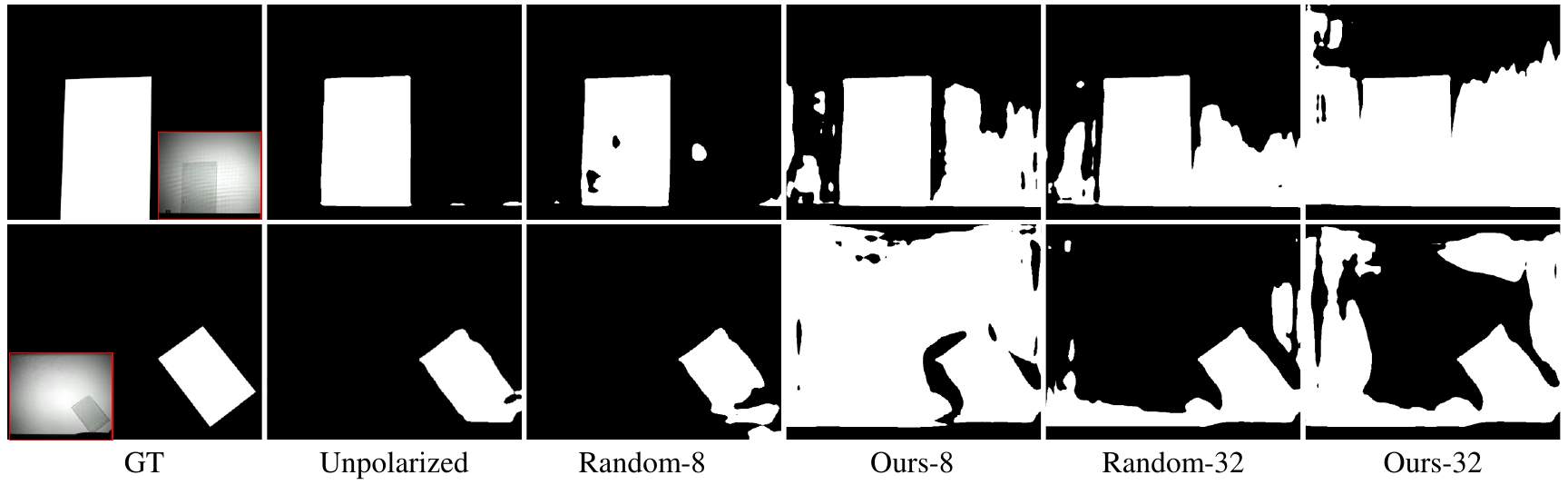}
	\caption{Visual comparison for adversarial attacks on the polarization-based glass segmentation model PGSNet~\cite{mei2022glass}. This comparison illustrates the predictions of inputs with projections in three states: unpolarized, polarized in random directions, and polarized in the optimized pattern using our technique. Random-$k$ and  Ours-$k$ represent random and our optimized perturbations at a grid size of $k$. The input images are visualized in red boxes.}
	\label{fig:glass}
\end{figure*}

\begin{figure}[t]
	\centering
	\includegraphics[width=1\linewidth]{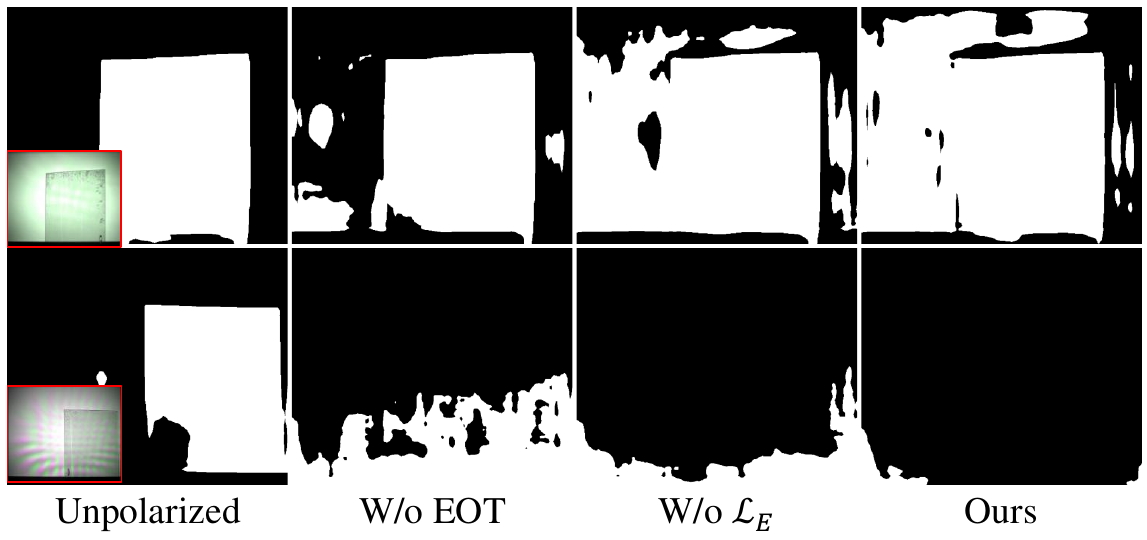}
	\caption{Visual comparison for ablation studies regarding EOT and $\mathcal{L}_E$, using optimized perturbations at a grid size of 8. }
	\label{fig:glassab}
	\vspace{-0.5cm}
\end{figure}

\begin{table}[t]
	\caption{Quantitative comparison using MAE and IoU in the digital and physical world. Random refers to perturbations that are randomly sampled, while Ours-$k$ denote our optimized perturbation at a grid size of $k$.}
	\label{table:a}
	\centering
	\begin{tabular}{l|cc|cc}
		\hline
		&\multicolumn{2}{|c|}{Digital world}  & \multicolumn{2}{c}{Physical world} 
		\\
		\hline
		& MAE$\uparrow$ & IoU $\downarrow$ &  MAE$\uparrow$& IoU$\downarrow$ 
		\\
		\hline
		
		Unpolarized  & 0.101&	0.715	&0.101&	0.715

		\\
		\hline
		Random-8 & 0.204&	0.461&
 	0.167&	0.509
		
		\\
		Ours-8 w/o EOT  & \bf{0.745}&	\bf{0.185}&	0.321&	0.411

		\\
		Ours-8 w/o $\mathcal{L}_E$&	0.569&	0.245	&0.271&	0.408

		\\
		Ours-8&	0.719&	0.200	&0.318&	0.422

		\\
		\hline
		Random-16 &0.246&	0.489
& 0.287&	0.449

		\\
		Ours-16 w/o EOT  & 0.735&	0.197&	0.372&	0.413

		\\
		Ours-16 w/o $\mathcal{L}_E$&	0.588&	0.205&	\bf{0.404}	& 0.396

		\\
		Ours-16&0.698&	0.190&	0.399&	0.377

		\\
		\hline
		Random-32 & 0.329&	0.440&	0.257&	0.474

		\\
		Ours-32 w/o EOT  & 0.674&	0.212&	0.307&	0.449

		\\
		Ours-32 w/o $\mathcal{L}_E$&0.680&	0.254&	0.342&	0.367

		\\
		Ours-32&	0.678&	0.250&	0.347&	\bf{0.361}

		\\
		\hline
	\end{tabular}
\end{table}
\vspace{-0.2cm}

\subsection{Experiments}
To simplify our experiments, we use a specific setup with a co-located projector and camera. This configuration obviates the need for calibrating their relative poses, thereby facilitating an effortless alignment of the camera's view with the projection. We gather candidate images and background images in an indoor setting. In Figure \ref{fig:sim}, we show the visual comparison between the digital world simulation and real-world captures. The AoLP and DoLP images show that our simulation approach reconstructs realistic polarization reflections at extremely high precision. On the contrast, the modification of rgb images from adversarial perturbation is visually imperceptible, thus realize an undermined adversarial attack on polarization-based vision model.

Experiments are conducted across 11 scenes to validate the efficacy of the proposed method. At an image resolution of $612\times512$, we set the grid size for our perturbations to be 8, 16, and 32, respectively, that a smaller grid size yields a higher resolution for the perturbation.  We apply MAE (Mean Absolute Error) and IoU (Intersection over Union) to characterize the prediction performance of the target model. For every scene, we run 300 iteration of optimizations, and $\lambda=1$ for $\mathcal{L}_E$. The step size is set to $\alpha=1$. We set $\tau$ as $0.3$ and reduce it gradually for an integration close to the candidate values. For a clear observation, the updates of $\mathcal{L}_{BCE}$ and IoU with different optimization settings are shown in Figure~\ref{fig:curve}. The visual comparison shows the $\mathcal{L}_{E}$ effectively enforce erroneous predictions especially with high resolution perturbations (grid size 8).

Quantitative evaluations are summarized in Table \ref{table:a}. 
Based on the results, our polarizing perturbations can significantly weaken the accuracy of PGSNet. The comparison reveals that in the digital world, perturbations with higher resolution are more effective in misleading the targeted model, especially in scenarios without the application of EOT. Conversely, in the physical world, adversarial perturbations with a lower resolution (grid size 32) exhibit superior performance. We attribute the improvement to the inherent robustness provided by grid-based perturbation. This robustness ensures effective transferability, even in the face of minor degradations. Furthermore, the use of (EOT) prevents overfitting to the input data and enhances the transferability of adversarial examples to real-world scenarios. Notably, our optimized adversarial examples outperform randomly generated perturbations with a great margin, and the use of our proposed $\mathcal{L}_E$ loss further amplifies attack efficacy in both digital and physical realms. 

We illustrate results of adversarial examples in two grid sizes, 8 and 32, as shown in Figure \ref{fig:glass}. 
When compared with predictions derived from inputs illuminated by unpolarized projection, the results highlight the efficacy of our polarizing projection in undermining the performance of PGSNet in both the digital and physical worlds. Notably, even perturbations that are randomly sampled can degenerate the model's performance.
Furthermore, perturbations synthesized via our optimization technique consistently outperform random perturbations. Particularly in physical world attacks, our method benefits substantially from the integration of Expectation Over Transformation (EOT) and the $\mathcal{L}_E$ loss function, resulting in robust and pronounced attack performance. Although minimal visual textures are discernible to the human eye, the polarization properties undergo significant alterations due to our perturbation projections. This approach effectively achieves both stealthiness and attack efficacy. Additional experimental details are provided in our supplementary material.

Visual comparisons of physical world attacks for the ablation study are also shown in Figure \ref{fig:glassab}, which indicate the significance of the proposed technique. The applied EOT enhances the robustness of the projection perturbation effectively and proposed $\mathcal{L}_E$ further boosts the attack performance. 

%% file: sec/5_shape.tex
\section{Whitebox Attack on Deep SfP}

\begin{figure*}[t]
	\centering
	\includegraphics[width=.95\linewidth]{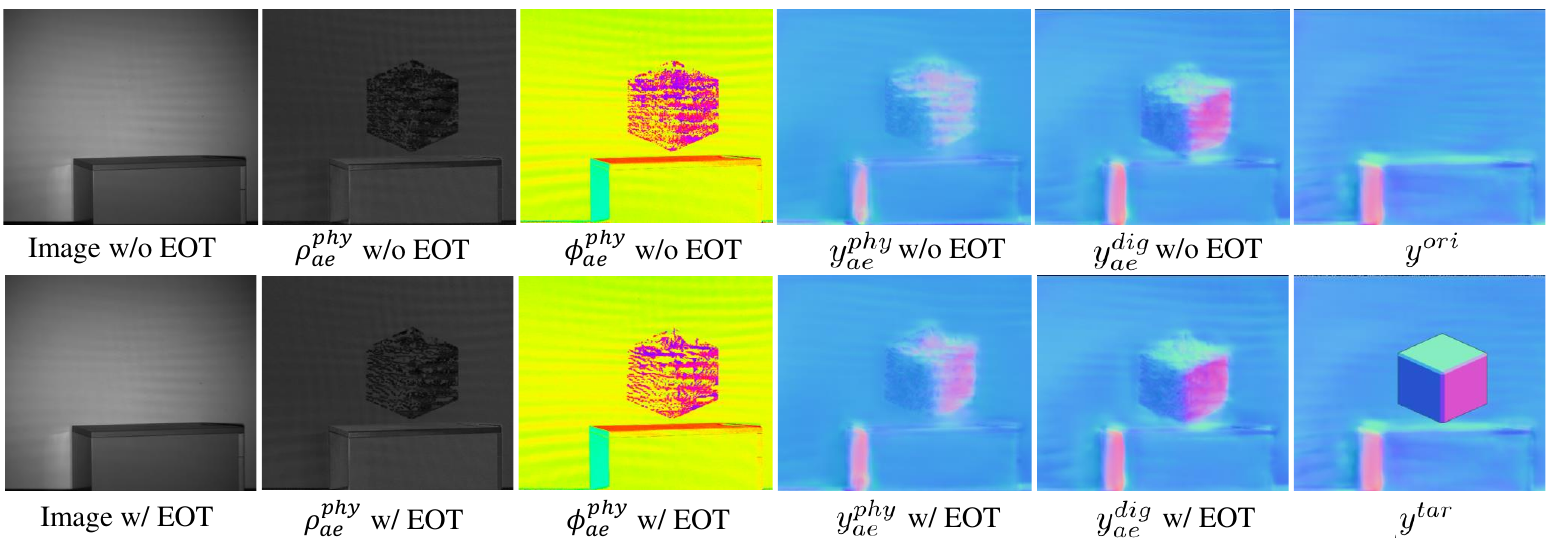}
	\vspace{-0.3cm}
	\caption{Visual comparison for adversarial attacks on the shape esimation model SfP-wild~\cite{lei2022shape}.}
	\label{fig:shape}
	\vspace{-0.5cm}
\end{figure*}

We have expanded our locally controllable polarizing projection technique to another key area of polarization imaging: shape estimation. Polarization imaging is inherently adept at capturing cues related to object geometries. In line with this, SfP-wild~\cite{lei2022shape} suggests leveraging deep priors from a large-scale polarization image dataset to estimate normal maps from single-shot images taken in the wild. Compared to RGB-only-based normal estimation, Lei's model derives significant advantages from polarization cues, enabling it to discern false geometries, e.g., the scene printed on a paper. Importantly, with deep priors derived from a large dataset captured in the wild, the model circumvents the lighting constraints~\cite{deschaintre2021deep,hwang2022sparse}. However, our experiments demonstrate instability when relying on polarization in the wild, that the perturbations imperceptible to the human eye will greatly influence the estimation results.

Our objective is to project a polarizing perturbation onto the background, aiming to deceive the model into estimating the shapes of non-existent 'objects'. Starting with an original estimation under a uniformly linear polarized projection, we superimpose the background normal map with an object, such as a cube, to serve as our target for the attack. We follow the settings described in Section 4 to optimize the perturbation pattern within the target region. For the optimization process, we simply employ the MAE (Mean Angular Error) loss~\cite{lei2022shape} and update the perturbation by  gradient descent. We introduce a high-resolution perturbation with a grid size of 2 and also incorporate the EOT methodology~\cite{athalye2018synthesizing} to ensure an effective adversarial perturbation in real world attacks.

In Figure \ref{fig:shape}, we show the adversarial examples for the physical world attacks, alongside the network outputs $y_{ae}^{l\in\{phy, dig\}}$ corresponding to physical and digital world attacks. Here, $y^{ori}$ represents the estimation obtained with the background illuminated by linear polarized projection, while $y^{tar}$ denotes the label of our targeted adversarial attack. As indicated by the intensity, $\rho^{phy}_{ae}$ and $\phi_{ae}^{phy}$ of adversarial examples, our perturbation focuses on modifying the polarization within the target region. In the digital world, the polarization domain modification is proved sufficient for misleading the network to generate a detailed normal map. While certain inherent challenges, such as noisy and quantized signals, the real-world attack leveraging EOT still yields results that closely align with the simulations. Please refer to our supplementary material for more evaluation.

In addition to the aforementioned experiments, we also conducted tests targeting the polarization-based color constancy algorithm~\cite{ono2022} and human pose and shape estimation model~\cite{zou2022human}. Further details on these experiments are provided in our supplementary material.

%% file: sec/6_conclusion.tex
\section{Research Ethics and Limitations}
This study originates from our curiosity on the potential vulnerability of polarization-based vision algorithms in the digital space. In line with existing researches on adversarial attacks, this study is intended to offer a timely warning on the potential vulnerability of polarization-based AI. 

The most obvious limitation we found lies in the relative low luminance of the projector, and the attack success rate will be low in bright environment. However, it is highly effective in indoor or low-light outdoor scenarios. Further protection measures, such as adversarial training, data enhancement, or introducing activate illuminations should be considered. 

\section{Conclusion}
Polarization has been utilized for a variety of computer vision tasks. We have shown that, similar to the well-known vulnerability of RGB-based vision, the performance of polarization-based vision algorithms, such as glass segmentation and shape estimation, can be manipulated, maybe in a potentially harmful way. Our adversarial attackers are physically realized by using an adapted one-chip LCD projector, which allows locally controllable polarizing projection. Our method is visually friendly, thus poses realistic concerns on the reliability of polarization-based AI. We hope this study will arouse attentions on the potential risks of polarization-based vision.

\noindent\textbf{Acknowledgement} This research was supported in part by JSPS KAKENHI Grant Numbers 22H00529, 20H05951, 23H03420, JST-Mirai Program JPMJMI23G1, and ROIS NII Open Collaborative Research 2023-23S1201.  